\documentclass[10pt,journal]{IEEEtran}
\IEEEoverridecommandlockouts
\usepackage{fancyhdr}
\usepackage{multirow}
\usepackage{mdwlist}
\usepackage{amsmath}
\usepackage{amssymb}
\usepackage{latexsym}
\usepackage{CJK}
\usepackage{subfigure}
\usepackage{indentfirst}
\usepackage{url}
\usepackage{graphicx}
\usepackage{subfigure}
\usepackage{epstopdf}
\usepackage{longtable}
\usepackage{array}
\usepackage[thmmarks,amsmath]{ntheorem}
\usepackage{diagbox}[2011/11/22]
\usepackage{textcomp,booktabs}
\usepackage[usenames,dvipsnames]{color}
\usepackage{colortbl}
\usepackage{breqn}
\usepackage{flushend}
\usepackage{mathrsfs}
\usepackage{stfloats}
\usepackage{algorithmic}
\usepackage{algorithm}
\usepackage{color}
\usepackage{bm}
\usepackage{engord}
\usepackage{makecell}
\usepackage{tcolorbox}

\usepackage{balance}
\usepackage{flushend}
\begin{document}

\title{\huge Joint Optimization of Model Partitioning and Resource Allocation for Anti-Jamming Collaborative Inference Systems}

\author{
Mengru Wu, Jiawei Li, Jiaqi Wei, Bin Lyu, Kai-Kit Wong, \IEEEmembership{Fellow, IEEE}, and Hyundong Shin, \IEEEmembership{Fellow, IEEE}
\vspace{-2em}

\thanks{\vspace*{-1em}}
\thanks{Mengru Wu, Jiawei Li, and Jiaqi Wei are with the College of Information Engineering, Zhejiang University of Technology, Hangzhou 310023, China (e-mail: wumengru@zjut.edu.cn; ljwjiawei@foxmail.com; wjqjiaqi@foxmail.com). \emph{(Corresponding author: Bin Lyu.)}}
\thanks{Bin Lyu is with the School of Communications and Information Engineering, Nanjing University of Posts and Telecommunications, Nanjing 210003, China (e-mail: blyu@njupt.edu.cn).}
\thanks{Kai-Kit Wong is with the Department of Electronic and Electrical Engineering, University College London, WC1E 7JE London, U.K., and also with the Department of Electronic Engineering, Kyung Hee University, Yongin-si, Gyeonggi-do 17104, Republic of Korea (e-mail: kai-kit.wong@ucl.ac.uk).}
\thanks{Hyundong Shin is with the Department of Electronics and Information Convergence Engineering, Kyung Hee University, Yongin-si, Gyeonggi-do 17104, Republic of Korea (e-mail: hshin@khu.ac.kr).}
}

\maketitle
\vspace{-1em}
\begin{abstract}
With the increasing computational demands of deep neural network (DNN) inference on resource-constrained devices, DNN partitioning-based device-edge collaborative inference has emerged as a promising paradigm.
However, the transmission of intermediate feature data is vulnerable to malicious jamming, which significantly degrades the overall inference performance.
To counter this threat, this letter focuses on an anti-jamming collaborative inference system in the presence of a malicious jammer. In this system, a DNN model is partitioned into two distinct segments, which are executed by wireless devices and edge servers, respectively.
We first analyze the effects of jamming and DNN partitioning on inference accuracy via data regression.
Based on this, our objective is to maximize the system's revenue of delay and accuracy (RDA) under inference accuracy and computing resource constraints by jointly optimizing computation resource allocation, devices' transmit power, and DNN partitioning.
To address the mixed-integer nonlinear programming problem, we propose an efficient alternating optimization-based algorithm, which decomposes the problem into three subproblems that are solved via Karush-Kuhn-Tucker conditions, convex optimization methods, and a quantum genetic algorithm, respectively.
Extensive simulations demonstrate that our proposed scheme outperforms baselines in terms of RDA.
\end{abstract}



\vspace{-0.5em}
\section{Introduction}
As a pivotal driving force in the evolution of sixth-generation (6G) communication networks, deep neural networks (DNNs) have been extensively deployed in various applications, such as object detection [1] and facial recognition [2]. However, the substantial computational demands of DNN inference present a performance bottleneck on resource-constrained devices that restricts inference efficiency. To address these challenges, DNN partitioning-based device-edge collaborative inference (CI) has emerged as a promising paradigm [3]. This approach involves partitioning a DNN model into two distinct segments and coordinating computation workloads between wireless devices and edge servers (ESs), thereby effectively enhancing resource utilization and reducing inference delay.

Driven by the rapid advancements in DNN inference, extensive research has focused on enabling DNN partitioning-based device-edge CI. Considering computation resource constraints, an adaptive DNN partitioning and resource allocation strategy was investigated in [4] to minimize onboard energy consumption while ensuring task completion. To accelerate DNN inference, an edge intelligence-empowered vehicular network was studied in [5], where DNN service placement and model partitioning were jointly optimized to minimize inference delay. Also, a robust DNN partitioning-based device-edge CI framework was developed in [6] to minimize the expected energy consumption of mobile devices under delay constraints. Integrating non-orthogonal multiple access with CI, an effective communication and computing resource allocation algorithm was introduced in [7] to achieve an optimal trade-off between energy consumption and inference delay.

Although the above studies [4]-[7] have explored multi-dimensional optimization for device-edge CI systems, they have failed to consider the threat of malicious jamming on the transmission of intermediate feature data (IFD).
In fact, malicious jamming over wireless links impairs the integrity of IFD, which further leads to significant degradation of inference accuracy [8].
Additionally, a DNN partitioning decision determines the size of IFD that needs to be transmitted, thus affecting the system’s susceptibility to jamming.
Motivated by the above discussions, this letter presents a joint design of model partitioning and resource allocation to achieve efficient collaboration against jamming in a device-edge CI system.
To the best of our knowledge, this is the first work to explore the impacts of DNN partitioning and transmission environment under malicious jamming on inference performance within a device-edge CI system.
To summarize, the main contributions of this letter are listed as follows:

\vspace{-0.2em}
\begin{itemize}
    \item
    We explore an anti-jamming device-edge CI system wherein multiple devices offload IFD to an ES in the presence of a malicious jammer.
    To characterize the effects of jamming and DNN partitioning, we adopt a data regression approach that relates inference accuracy to signal-to-interference-plus-noise ratio (SINR) and partitioning point via a logistic function.

    \item
    We introduce a revenue of delay and accuracy (RDA) metric to evaluate system performance under constraints on inference accuracy and computing resources.
    Based on this, we formulate an optimization problem to maximize the system's RDA by jointly optimizing computation resource allocation, devices' transmit power, and DNN~partitioning.


    \item
    To tackle this non-convex problem, we propose an efficient iterative algorithm that decomposes it into three subproblems.
    We first derive a closed-form solution for computation resource allocation via the Karush-Kuhn-Tucker (KKT) conditions.
    Then, we employ convex optimization methods to determine each device's transmit power.
    To solve the DNN partitioning subproblem, we adopt a quantum genetic algorithm (QGA).
    Extensive simulations validate the superiority of our scheme over~baselines.

\end{itemize}

\section{System Model and Problem Formulation}
\subsection{System Model}
As illustrated in Fig. 1, we consider a device-edge CI system consisting of an ES, $N$ wireless devices indexed by the set ${\cal N} = \left\{ 1,2, \ldots, N\right\}$, and a malicious jammer. In this network, each device performs a computation-intensive inference task using a DNN model. Considering the resource limitations on devices, we partition the DNN model to distribute the computation load effectively across devices and the ES. Specifically, device $n \in {\cal N}$ executes the initial layers of the DNN model to handle its inference task and then transmits the extracted IFD to the ES for further processing using the remaining layers of the DNN model. In addition, a malicious jammer sends jamming signals to impair the wireless link between devices and the ES.
Due to inherent channel noise and malicious jamming, the degradation of IFD quality during transmission compromises inference accuracy.

\begin{figure}[t]
\centering
\setlength{\abovecaptionskip}{0.cm}
\setlength{\belowcaptionskip}{-0.cm}
\includegraphics[height=32mm,width=84mm]{{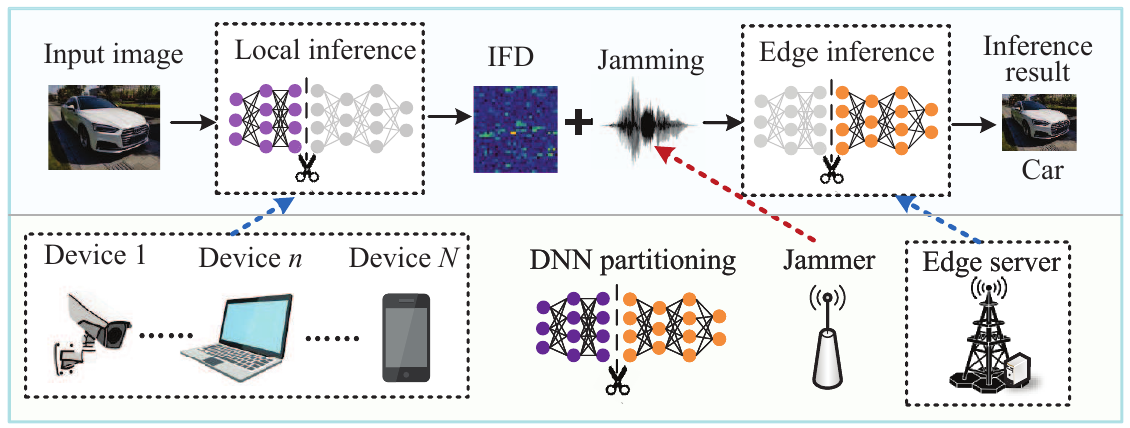}}
\vspace{-0.5em}
\caption{A device-edge CI system.}
\vspace{-0.5em}
\end{figure}

\subsubsection{DNN Partitioning-based CI}
The DNN model consists of multiple layers, which can be represented by ${\cal K} = \left\{ 0, 1, 2, \ldots, K \right\}$.
We denote $k_n \in {\cal K}$ as the partitioning point of the DNN~model at device $n$, which means that device $n$ executes the layers from $0$ to $k_n$, while the ES is responsible for the subsequent layers from $k_n+1$ to $K$.
In particular, $k_n=0$ indicates that device $n$'s inference task is entirely offloaded to the ES, while $k_n=K$ signifies that device $n$ performs the full inference task locally.
Thus, the total computation workloads of device $n$ and the ES can be respectively expressed as
\vspace{-0.5em}
\begin{equation}
L_n^{D}\left(k_n\right) = \sum_{k=0}^{k_n} l_n\left(k\right), {\kern 10pt} L_n^{E}\left(k_n\right) = \sum_{k=k_n+1}^{K}\!\!\!\!l_n\left(k\right),
\vspace{-0.5em}
\end{equation}
where $l_n\left(k\right)$ is the computation workload of layer $k$ for device $n$'s task.
Then, the local inference delay and energy consumption of device $n$ can be respectively written as
\vspace{-0.5em}
\begin{equation}
t_n^D = {{L_n^D\left( {{k_n}} \right)} \mathord{\left/{\vphantom {{L_n^D\left( {{k_n}} \right)} {{f_n}}}} \right.\kern-\nulldelimiterspace} {{f_n}}},
\end{equation}
\vspace{-1em}
\begin{equation}
\Xi _n^{D}=\zeta L_n^D\left({k_n}\right){\left( f_n \right)^2},
\vspace{-0.5em}
\end{equation}
where $f_n$ represents device $n$'s local computation capability and $\zeta$ means a coefficient related to the chip architecture.
Besides, the corresponding inference delay of the ES is given~by
\vspace{-0.5em}
\begin{equation}
t_n^E = {{L_n^E\left( {{k_n}} \right)} \mathord{\left/{\vphantom {{L_n^E\left( {{k_n}} \right)} {{f_n^E}}}} \right.\kern-\nulldelimiterspace} {{f_n^E}}},
\vspace{-0.5em}
\end{equation}
where $f_n^E$ equals the computation resources allocated to process device $n$'s task.

\subsubsection{IFD Transmission}
To avoid mutual interference among devices, orthogonal frequency resources are adopted for IFD transmission between devices and the ES.
However, the jammer generates intentional jamming signals, which significantly degrade the quality of the transmitted IFD.
On this basis, the SINR between device $n$ and the ES can be expressed as
\vspace{-0.5em}
\begin{equation}
\phi _n = \frac{{p_n h_n}}{{{p_j}{h_j} + \sigma^2}},
\vspace{-0.5em}
\end{equation}
where $p_n$ and $p_j$ represent the transmit power of device $n$ and the jammer, respectively,
$h_n$ and $h_j$ denote the channel coefficients from device $n$ and the jammer to the ES, and $\sigma^2$ refers to the noise power.
Thus, the transmission delay and energy consumption of device $n$ for sending IFD are respectively derived as
\vspace{-0.5em}
\begin{equation}
t_n^{T} = \frac{{S_n \left({k_n}\right)}}{{B_n}{\log _2}\left( {1 +\phi _n} \right)},
\end{equation}
\vspace{-0.5em}
\begin{equation}
\Xi _n^T = \frac{{{p_n}{S_n}\left({k_n}\right)}}{{B_n}{\log _2}\left( {1 +\phi _n} \right)},
\vspace{-0.5em}
\end{equation}
where $B_n$ is the transmission bandwidth and $S_n \left({k_n}\right)$ indicates the size of IFD extracted by the DNN partitioning layer $k_n$.

\subsubsection{Inference Accuracy Model}
In the CI system, the inference accuracy is significantly influenced by both the transmission quality of IFD and the model partitioning strategy. To be specific, the SINR of the wireless channel directly affects the integrity of data transmission. Meanwhile, the selection of the DNN partitioning point dictates the size of the IFD.
To quantify this joint impact, we derive a closed-form accuracy model via regression analysis.
Specifically, the inference accuracy can be obtained by conducting experiments across a range of partitioning points and SINRs, from which we derive the inference accuracy function as [9]
\vspace{-0.5em}
\begin{equation}
\overline{\rm{Acc}}_n \left({k_n},{\phi _n}\right) = \frac{{{A_{{k_n}}}}}{{1 + {e^{ - {\tau _{{k_n}}}\left({\phi _n} - {\phi _{{k_n}}}\right)}}}} + {b_{{k_n}}},
\vspace{-0.5em}
\end{equation}
where $A_{k_n}$, $\tau _{k_n}$, $\phi _{k_n}$, and $b_{k_n}$ are fitting parameters determined by conducting experiments.
Note that these fitting parameters can be adjusted for different DNN models or datasets, enabling applicability to other systems.

\vspace*{-1em}
\subsection{Problem Formulation}
Based on the above analysis, we respectively define the delay revenue and accuracy revenue for each device as
\vspace{-0.5em}
\begin{equation}
{\widehat T}_n= \frac{{{T_{\max }} - \left({t_n^D+t_n^T+t_n^E}\right)}}{{{T_{\max }}}},
\end{equation}
\vspace{-0.5em}
\begin{equation}
{\widehat {\rm Acc}}_n=\frac{{\overline {\rm Acc}}_n- {\rm Acc}_{\min}}{{\rm Acc}_{\max}-{\rm Acc}_{\min}},
\vspace{-0.5em}
\end{equation}
where $T_{\max}$ equals the maximum delay, ${\rm Acc}_{\max}$ represents the maximum achievable accuracy, and ${\rm{Acc}}_{\min}$ stands for the minimum inference accuracy threshold.

In this paper, our objective is to maximize the RDA for device-edge CI subject to inference accuracy requirements and computing resource constraints.
To achieve this goal, we propose a joint optimization framework that determines the computation resource allocation ${\bf F}=\left\{f_n^E, \forall n\in {\cal N}\right\}$, the transmit power of each device ${\bf P} = \left\{ {p_n},\forall n \in {\cal N}\right\}$, and the DNN partitioning ${\bf K} = \left\{ {k_n},\forall n \in {\cal N}\right\}$. Then, we formulate the system's RDA maximization problem as
\vspace{-0.5em}
\begin{subequations}
\begin{align}
{\mathcal P}_0\!:& \!\!{\kern 12pt}\max_{\bf{F}, \bf{P}, \bf{K}} \sum\limits_{n = 1}^N \left(\xi {\widehat T}_n + \left( {1 - \xi } \right){\widehat {\rm Acc}}_n\right){\kern 12pt} \\
\rm{s.t.} & {\kern 11pt} \overline{\rm{Acc}}_n \ge {\rm{Acc}}_{\min}, {\kern 8pt}\forall n \in {\cal N}, \\
 & {\kern 11pt} \Xi _n^D + \Xi _n^T \le {\Xi _{\max}},{\kern 8pt}\forall n \in {\cal N},\\
 &{\kern 10pt}  \sum_{n=1}^{N} f_n^{E} \le F_{\max},{\kern 8pt}\forall n \in {\cal N},\\
 & {\kern 11pt} 0 < {p_n} \le {P_{\max }},{\kern 8pt}\forall n \in {\cal N},\\
 & {\kern 11pt} {k_n} \in \left\{ 0,1,...,K\right\},{\kern 8pt}\forall n \in {\cal N},
 \vspace{-0.5em}
\end{align}
\end{subequations}
where $\xi $ is a positive weighting coefficient for the delay-accuracy trade-off, ${\Xi}_{\max}$ and $P_{\max}$ mean the maximum available energy and transmit power of each device, respectively, and $F_{\max}$ denotes the maximum computation capacity of the ES. Here, (11b) ensures that the accuracy of inferring each device's task meets the minimum accuracy threshold. (11c) implies that each device's energy consumption cannot exceed its maximum available energy. (11d) specifies that the computation resources allocated for task processing should not surpass the maximum computing ability of the ES. (11e) and (11f) are used to restrict the feasible regions of the optimization variables.

\section{Proposed Solution}
Due to the complex interplay between discrete and continuous variables, $\mathcal{P}_0$ is a mixed-integer nonlinear programming (MINLP) problem that is difficult to solve directly. To address this problem, we propose an efficient iterative algorithm based on the alternating optimization (AO) framework.
Specifically, we decouple the joint optimization into three subproblems that respectively address computation resource allocation, transmit power optimization, as well as DNN partitioning.
By iteratively solving these subproblems until convergence, we ultimately obtain a suboptimal solution.
Next, we provide details for solving each subproblem.

\vspace*{-1em}
\subsection{Computation Resource Allocation Subproblem}
Given the transmit power and DNN partitioning, we focus on the subproblem of computation resource allocation as
\vspace{-0.5em}
\begin{subequations}
\begin{align}
{\mathcal P}_1\!:& \!\!{\kern 12pt}\max_{\bf{F}} \sum\limits_{n = 1}^N \left(\xi {\widehat T}_n + \left( {1 - \xi } \right){\widehat {\rm Acc}}_n\right){\kern 12pt} \\
{\rm{s.t.}} & {\kern 11pt} \text{(11d).}
\vspace{-0.5em}
\end{align}
\end{subequations}
It is obvious that problem $\mathcal{P}_1$ is a convex optimization problem. To solve this problem, we employ the Lagrangian method for deriving the solution. We first present the expression of the Lagrangian function for $\mathcal{P}_1$ as
\vspace{-0.5em}
\begin{equation}
{\cal L}\!\left( {f_n^E,\lambda } \right) \!= \!\sum\limits_{n = 1}^N \!{ \left(-\xi {\widehat T}_n\!+\! \left( {\xi \!-\! 1} \right){\widehat {\rm Acc}}_n\right)} \!+\! \lambda \!\left( {\sum\limits_{n = 1}^N \!{f_n^E}\!-\!{F_{\max }}} \right),
\end{equation}
where $\lambda$ is a nonnegative Lagrange multiplier. Thus, the KKT conditions for optimality include
\vspace{-0.5em}
\begin{equation}
\frac{{\partial {\cal L}\left( {f_n^E,\lambda } \right)}}{{\partial f_n^E}} = \frac{{ - \xi L_n^E\left( {{k_n}} \right)}}{{{T_{\max}{\left( {f_n^E} \right)}^2}}} + \lambda  = 0,
\vspace{-0.5em}
\end{equation}
\begin{equation}
\lambda \left( {\sum\limits_{n = 1}^N {f_n^E}-{F_{\max }}}\right)=0.
\vspace{-0.5em}
\end{equation}
By jointly addressing (14) and (15), we can derive the optimal computation resource allocation solution as
\vspace{-0.5em}
\begin{equation}
f_n^{E*}= \frac{{{F_{\max }}\sqrt {L_n^E\left( {{k_n}} \right)} }}{{\sum\limits_{n = 1}^N {\sqrt {L_n^E\left( {{k_n}} \right)} } }}.
\vspace{-0.5em}
\end{equation}


\vspace*{-1em}
\subsection{Transmit Power Optimization Subproblem}
With given computation resource allocation and DNN partitioning, the transmit power subproblem can be expressed as
\vspace{-1.5em}
\begin{subequations}
\begin{align}
{\mathcal P}_2\!:& \!\!{\kern 12pt}\max_{\bf{P}} \sum\limits_{n = 1}^N \left(\xi {\widehat T}_n + \left( {1 - \xi } \right){\widehat {\rm Acc}}_n\right){\kern 12pt} \\
{\rm{s.t.}} & {\kern 11pt} \text{(11b), (11c), (11e).}
\vspace{-0.5em}
\end{align}
\end{subequations}
This problem is non-convex and cannot be addressed directly.
Given that the objective increases monotonically with transmit power, we reformulate it as maximizing the total transmit power.
Meanwhile, constraints (11b) and (11c) are equivalently transformed into convex forms as
\vspace{-0.5em}
\begin{equation}
{\phi _n} - {\phi _{{k_n}}} \ge  - \frac{1}{{{\tau _{{k_n}}}}}\ln \left( {\frac{{{A_{{k_n}}}}}{{{\rm{Ac}}{{\rm{c}}_{\min }} - {b_{{k_n}}}}} - 1} \right),
\vspace{-0.5em}
\end{equation}
\vspace{-0.5em}
\begin{equation}
\left( {{\Xi _{\max }} - \Xi _n^D} \right){B_n}{\log _2}\left( {1 +\phi _n} \right) - {p_n}{S_n}\left( {{k_n}} \right) \ge 0.
\vspace{-0.5em}
\end{equation}
Thus, problem $\mathcal{P}_2$ can be rewritten as
\vspace{-0.5em}
\begin{subequations}
\begin{align}
\mathcal{P}_2'\!:\!\! & {\kern 12pt}\max_{\bf{P}} \sum\limits_{n = 1}^N {p_n}{\kern 20pt}\\
{\rm{s.t.}} &{\kern 11pt} \text{(18), (19), (11e).}
\vspace{-0.5em}
\end{align}
\end{subequations}
Problem $\mathcal{P}_2'$ is a convex optimization problem that can be solved efficiently using standard convex solvers, such as CVX.

\vspace*{-1em}
\subsection{DNN Partitioning Subproblem}
With computation resource allocation and transmit power fixed, the DNN partitioning subproblem is written as
\vspace{-0.5em}
\begin{subequations}
\begin{align}
{\mathcal P}_3\!:& \!\!{\kern 12pt}\max_{\bf{K}} \sum\limits_{n = 1}^N \left(\xi {\widehat T}_n + \left( {1 - \xi } \right){\widehat {\rm Acc}}_n\right){\kern 12pt} \\
{\rm{s.t.}} & {\kern 11pt} \text{(11b), (11c), (11f).}
\vspace{-0.5em}
\end{align}
\end{subequations}
It is clear that $\mathcal{P}_3$ is an integer programming problem due to the discrete DNN partitioning variables.
To address this problem, we design a QGA-based DNN partitioning algorithm to search for a suboptimal solution.
QGA is a probabilistic optimization algorithm that integrates the principles of quantum computing with genetic algorithms (GAs), which effectively mitigates premature convergence and enhances global optimization performance [10].



In the QGA, each chromosome is encoded by qubits to characterize a probabilistic distribution of DNN partitioning solutions across $N$ devices, with each device's DNN partitioning decision represented by $m=\lceil \log_2(K+1) \rceil$ qubits.
The chromosome of the $i$-th individual in the $t$-th generation is given by
\vspace{-0.5em}
\begin{equation}
{\bf q}_i^{\left( t \right)} =
\begin{bmatrix}
\alpha_{i,1}^{\left( t \right)} &\cdots &\alpha_{i,j}^{\left( t \right)} &\cdots &\alpha_{i,J}^{\left( t \right)} \\
\beta_{i,1}^{\left( t \right)} &\cdots &\beta_{i,j}^{\left( t \right)} &\cdots &\beta_{i,J}^{\left( t \right)}
\end{bmatrix},
\vspace{-0.5em}
\end{equation}
where $\alpha_{i,j}^{\left( t \right)}$ and $\beta_{i,j}^{\left( t \right)}$ denote the probability amplitudes of the $j$-th qubit, which satisfy ${\left| \alpha_{i,j}^{\left( t \right)}  \right|^2}\!+\!{\left| \beta_{i,j}^{\left( t \right)}\right|^2}\!\!=\!1$, and $J\!=\!mN$ is the total number of~qubits.
The qubit-based chromosome exists in a superposition of multiple candidate solutions.
Upon measurement, the chromosome ${\bf q}_i^{\left( t \right)}$ collapses into a deterministic binary string as ${\bf x}_i^{\left( t \right)} = \left[ {x_{i,1}^{\left( t \right)}, \ldots ,x_{i,j}^{\left( t \right)}, \ldots ,x_{i,J}^{\left( t \right)}} \right]$, where each bit $x_{i,j}^{\left( t \right)}$ is obtained by
\vspace{-0.5em}
\begin{equation}
x_{i,j}^{\left( t \right)}=\left\{ {\begin{array}{*{20}{l}}
{1,}&{{\rm{if }} {\kern 4pt}r_{i,j} \le {\left|\beta _{i,j}^{\left( t \right)}\right|}^2,}\\
{0,}&{{\rm{otherwise.}}}
\end{array}} \right.
\vspace{-0.5em}
\end{equation}
Here, $r_{i,j}$ is a random number in $\left[0,1\right)$.
Then, we map the binary string ${\bf x}_i^{\left( t \right)}$ to the DNN partitioning solution ${\bf k}_i^{\left( t \right)} = \left[ {k_{i,1}^{\left( t \right)}, \ldots ,k_{i,n}^{\left( t \right)}, \ldots ,k_{i,N}^{\left( t \right)}} \right]$, where each element $k_{i,n}^{\left( t \right)}$ satisfies
$k_{i,n}^{\left( t \right)} = \sum\limits_{j = 1}^m {{2^{m - j}}x_{i,\left( {n - 1} \right)m + j}^{\left( t \right)}}$.
To evaluate each individual, we define a fitness function as
\vspace{-0.5em}
\begin{equation}
\begin{split}
&Fit  \left({\bf k}_i^{\left( t \right)}\right)=\sum\limits_{n = 1}^N \left({\xi {\widehat T}_n + \left( {1- \xi } \right){\widehat {\rm Acc}}_n}\right. \\
&\!+\!{\omega _1}{\min \!{\left(0, \overline{\text{Acc}}_n\! -\! {\text{Acc}}_{\min}\right)}}\!+\!{\omega _2}{\min \!{\left(0, \Xi_{\max}\!-\!\Xi _n^D\!-\! \Xi _n^T \right)}}\!\Big),
\end{split}
\vspace{-0.5em}
\end{equation}
where $\omega_1$ and $\omega_2$ are penalty coefficients.
Based on the fitness evaluation, the quantum chromosomes are updated using quantum rotation gates to evolve the population toward better solutions.
The rotation operation for the $j$-th qubit of the $i$-th individual is performed as
\vspace{-0.5em}
\begin{equation}
\left[ {\begin{array}{*{20}{l}}
{\alpha _{i,j}^{\left( {t + 1} \right)}}\\
{\beta _{i,j}^{\left( {t + 1} \right)}}
\end{array}} \right] = \left[ {\begin{array}{*{20}{c}}
{\cos \left({\theta _{i,j}}\right)}&{ - \sin \left({\theta _{i,j}}\right)}\\
{\sin \left({\theta _{i,j}}\right)}&{\cos \left({\theta _{i,j}}\right)}
\end{array}} \right]\left[ {\begin{array}{*{20}{l}}
{\alpha _{i,j}^{\left( t \right)}}\\
{\beta _{i,j}^{\left( t \right)}}
\end{array}} \right],
\vspace{-0.5em}
\end{equation}
where $\alpha_{i,j}^{\left( t+1 \right)}$ and $\beta_{i,j}^{\left( t+1 \right)}$ represent the updated state of the qubit, and $\theta_{i,j}$ is a rotation angle.
To enhance individual diversity, the quantum crossover is conducted with a specific crossover probability $p_c$, allowing for the exchange of quantum information between randomly selected individuals.
Additionally, with a mutation probability $p_m$, the quantum mutation is applied to randomly flip the state of selected qubits.
Based on this, the quantum chromosomes are iteratively updated to search for the optimal solution until convergence.

\begin{algorithm}[t]
\caption{AO-based Algorithm for Solving $\mathcal {P}_0$}
\begin{algorithmic}[1]
\setlength{\belowcaptionskip}{-1cm}
\STATE Set feasible values of ${\bf P}^{\left(0\right)}$, ${\bf K}^{\left(0\right)}$, and ${\bf F}^{\left(0\right)}$ and ${\tilde t}= 1$.
\REPEAT
    \STATE Given ${\bf P}^{\left( {\tilde t}-1\right)}$ and ${\bf K}^{\left( {\tilde t}-1\right)}$, solve problem $\mathcal{P}_1$  according to (16) and obtain ${\bf F}^{\left({\tilde t}\right)}$.
    \STATE Given ${\bf K }^{\left({\tilde t}-1\right)}$ and ${\bf F}^{\left({\tilde t}\right)}$, solve problem $\mathcal{P}_2$  using CVX and obtain ${\bf P}^{\left( {\tilde t}\right)}$.
    \STATE Given ${\bf P}^{\left( {\tilde t}\right)}$ and ${\bf F}^{\left({\tilde t}\right)}$, solve problem $\mathcal{P}_3$  using QGA and obtain ${\bf K}^{\left({\tilde t}\right)}$.
    \STATE Update ${\tilde t} \leftarrow {\tilde t}+1$.
\UNTIL convergence
\end{algorithmic}
\end{algorithm}

Consequently, the proposed AO-based algorithm iteratively optimizes computation resource allocation, devices' transmit power, and DNN partitioning until convergence. The overall algorithm is summarized in Algorithm 1.
Also, the overall computational complexity of Algorithm 1 is expressed as $\mathcal {O}\left({T_{\rm{AO}}}\left({N} +{N^{3.5}} + J{N_p}{t_{\max}}\right)\right)$, where ${T_{\rm{AO}}}$ represents the number of AO iterations, $N_p$ equals the population size, and $t_{\max}$ denotes the maximum number of generations.

\section{Simulation Results}
In this section, we perform simulations to evaluate the performance of the proposed scheme.
We consider a CI scenario with ten devices, an ES, and a malicious jammer, all randomly distributed in a square region of $100\times100$ ${\rm m}^2$.
The channel gains follow $h_i=10^{-3}d_i^{-3}$, $i\in \left\{n,j\right\}$, where $d_n$ and $d_j$ are the distances from device $n$ and the jammer to the ES, respectively. Unless otherwise specified, the simulation parameters are set according to [11]. To be specific, we consider $f_n = 2$ GHz, $F_{\max} = 20$ GHz, $B_n = 1$ MHz, $P_{\max} = 0.23$ W, $p_j=1$ W, $\sigma ^2=-110$ dBm, $\zeta =10^{-28}$, $\Xi_{\max}=1$ J, $\xi=0.5$, $K=5$, ${\rm{Acc}}_{\min}=80$ \%, ${\rm{Acc}}_{\max}=95$~\%, and $T_{\max}=2$~s.
In the QGA, the parameters are configured as $N_p = 100$, $t_{\max}=100$, $p_c=0.8$, $p_m=0.02$, $\omega _1 = 10^6$, and $\omega _2= 10^6$.
Also, we set $\theta_{i,j}=-0.01\pi$ if the $j$-th observed bits of the best and the $i$-th individuals are equal, and $\theta_{i,j}=0.05\pi$ otherwise.

In addition, we adopt a ResNet 18 model trained on the CIFAR 10 dataset to characterize the DNN inference task, as shown in Fig. 2.
Based on the above experimental setup, we obtain the fitting parameters presented in Fig. 3.
Specifically, the corresponding fitting parameter sets for $A_{k_n}$, $\tau_{k_n}$, $\phi_{k_n}$, and $b_{k_n}$ across different partitioning points $k_n$ are $\{0.86,$ $0.85,$ $0.83,$ $0.89,$ $0.84\}$, $\{0.38,$ $0.70,$ $0.46,$ $0.42,$ $0.57\}$, $\{6.98,$ $7.29,$ $11.79,$ $14.08,$ $11.58\}$, and $\{0.09,$ $0.10,$ $0.12,$ $0.06,$ $0.11\}$,~respectively.

To clarify the effectiveness of the proposed scheme, we compare it with the following baselines:
1) Local computing (LC) scheme where all tasks are processed locally on devices.
2) Edge server computing (ESC) scheme that offloads all tasks to the ES for processing.
3) Fixed transmit power (FTP) scheme in which each device transmits its extracted IFD to the ES using a fixed transmit power.
4) GA-based scheme that utilizes a GA for DNN partitioning [12], while other variables are determined by our proposed algorithm.

\begin{figure}[t]
\centering
\setlength{\abovecaptionskip}{0.cm}
\setlength{\belowcaptionskip}{-0.cm}
\includegraphics[height=18mm,width=90mm]{{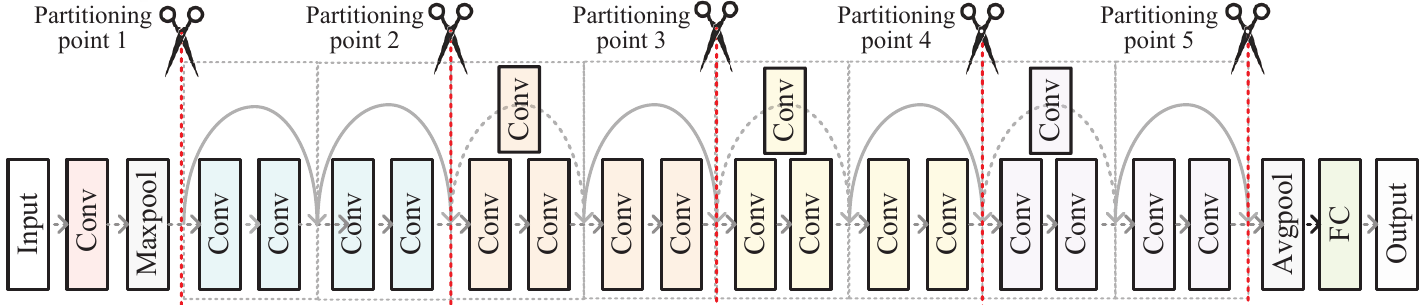}}
\caption{Partitioning points of ResNet 18.}
\end{figure}

\begin{figure}[t]
\centering
\setlength{\abovecaptionskip}{0.cm}
\setlength{\belowcaptionskip}{-0.cm}
\includegraphics[height=36mm,width=43mm]{{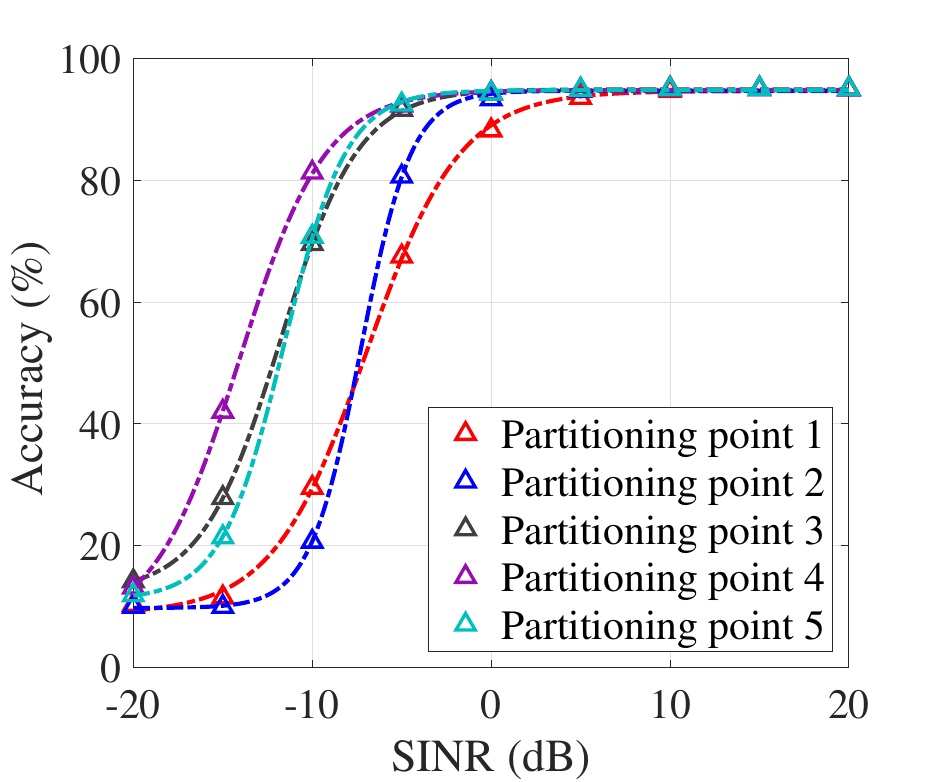}}
\vspace{-0.5em}
\caption{Data fitting of ResNet 18 on the CIFAR 10 dataset.}
\vspace{-0.5em}
\end{figure}

\begin{figure*}[!t]
\setlength{\abovecaptionskip}{0 cm}
\setlength{\belowcaptionskip}{-0.2 cm}
\centering
\subfigure[]{\raisebox{-1ex}{\includegraphics[height=38mm,width=48mm]{{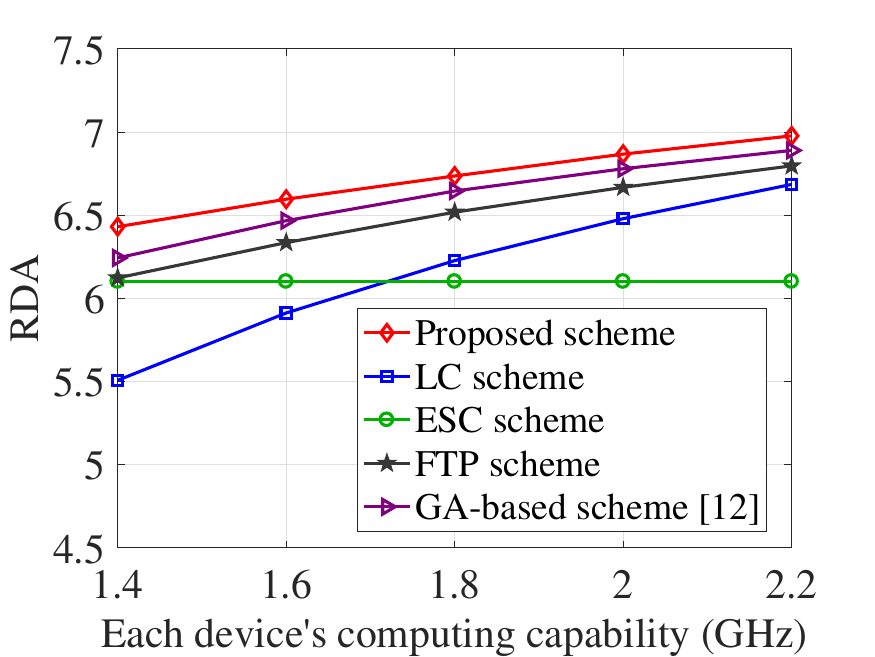}}}}
\subfigure[]{\raisebox{-1ex}{\includegraphics[height=38mm,width=48mm]{{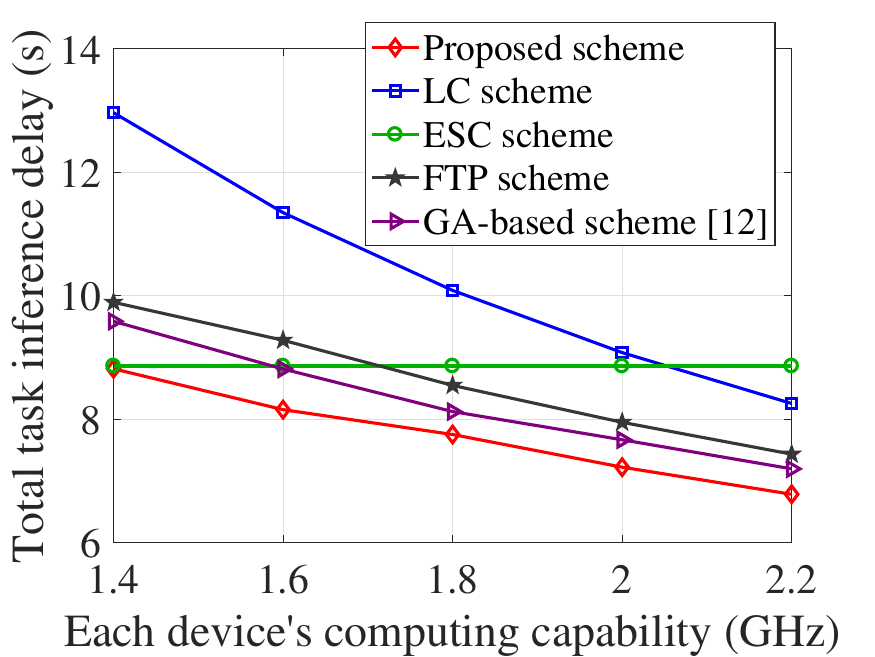}}}}
\subfigure[]{\raisebox{-1ex}{\includegraphics[height=38mm,width=48mm]{{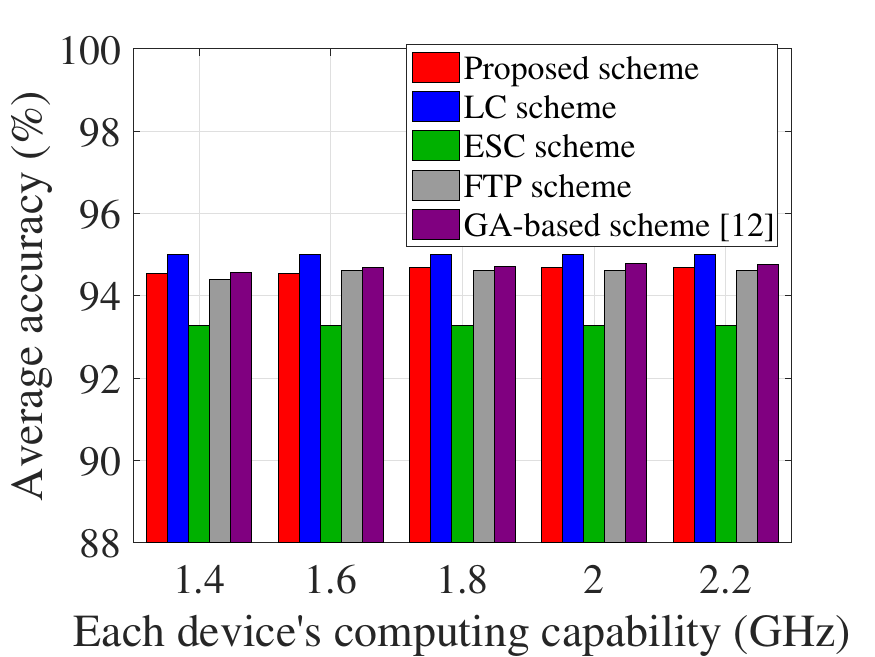}}}}
\vspace{-0.3em}
\caption{Performance with respect to each device's computing capability: (a) RDA. (b) Total task inference delay. (c) Average accuracy.}
\vspace{-1em}
\end{figure*}

\begin{figure*}[t!]
\setlength{\abovecaptionskip}{0 cm}
\setlength{\belowcaptionskip}{-0.2 cm}
\centering
\subfigure[]{\raisebox{-1ex}{\includegraphics[height=38mm,width=48mm]{{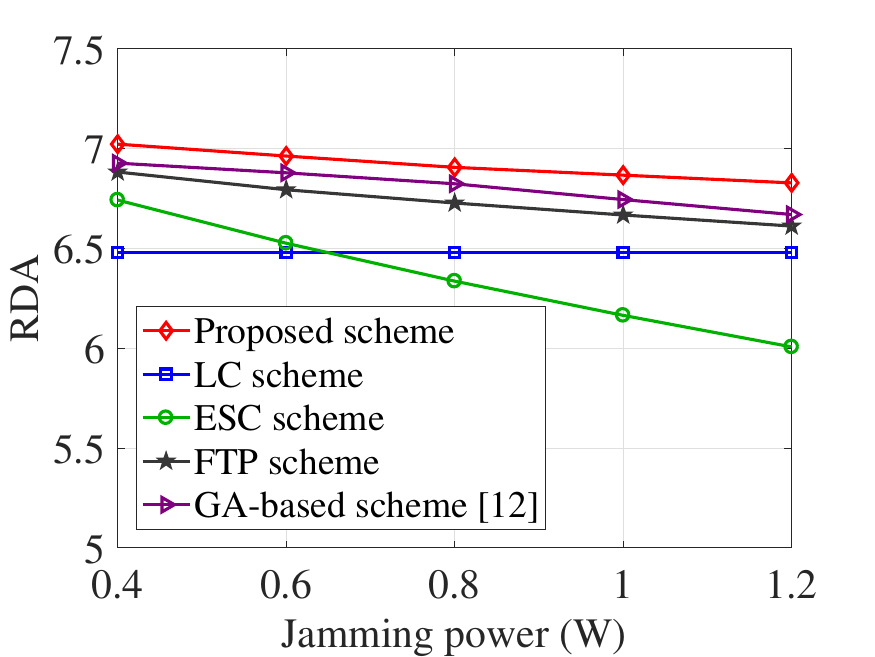}}}}
\subfigure[]{\raisebox{-1ex}{\includegraphics[height=38mm,width=48mm]{{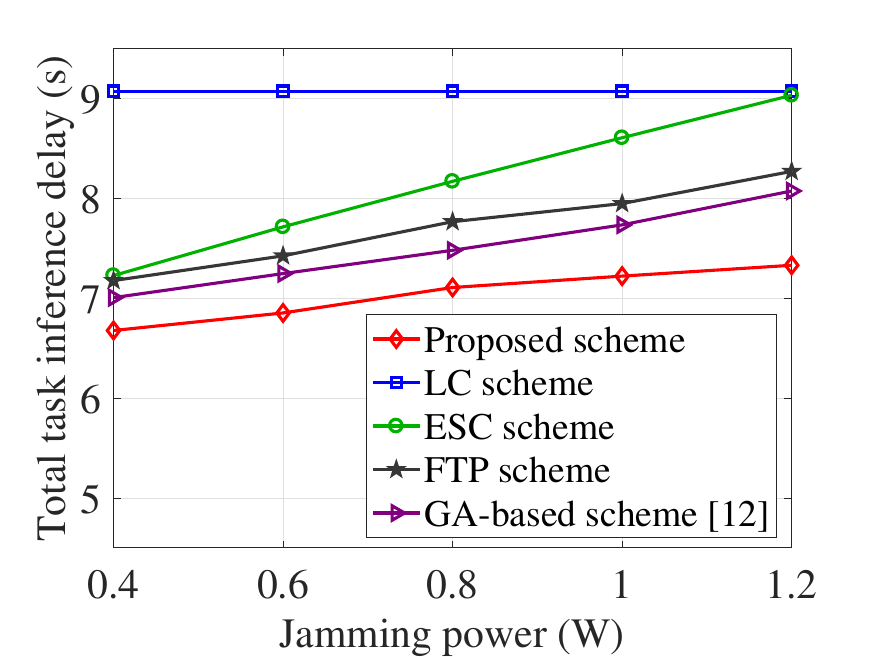}}}}
\subfigure[]{\raisebox{-1ex}{\includegraphics[height=38mm,width=48mm]{{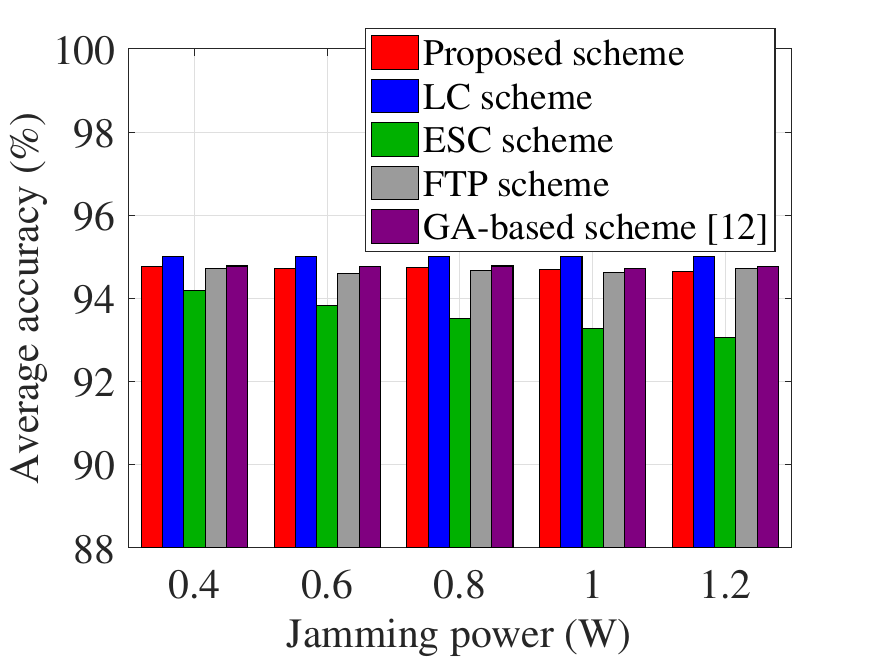}}}}
\vspace{-0.3em}
\caption{Performance with respect to jamming power: (a) RDA. (b) Total task inference delay. (c) Average accuracy.}
\vspace{-1em}
\end{figure*}

In Figs. 4(a)$\sim$(c), we access the performance in terms of RDA, total task inference delay, and average accuracy against each device's computing capability.
As shown in Fig. 4(a), the proposed scheme achieves the highest RDA among baselines. This demonstrates the superiority of our joint optimization framework in effectively balancing delay and inference accuracy under jamming.
In Fig. 4(b), as the computing capability of each device increases, the total task inference delay decreases across all schemes except the ESC scheme. This is because improved local computing capabilities allow devices to process a larger proportion of tasks locally.
In Fig. 4(c), although the proposed scheme does not achieve the highest accuracy, it realizes the best trade-off between delay and inference accuracy while satisfying the inference accuracy~threshold.

We present the RDA, total task inference delay, and average accuracy versus jamming power in Figs. 5(a)$\sim$(c).
In Fig. 5(a), the proposed scheme consistently outperforms all baselines in terms of RDA across varying jamming power, demonstrating its robust anti-jamming capability.
Fig. 5(b) shows that total task inference delay increases with jamming power for all schemes except the LC scheme, as transmission quality degradation drives devices to adjust transmit power and partitioning strategies.
In Fig. 5(c), the proposed scheme maintains high inference accuracy above the required threshold even under strong jamming, highlighting its reliability and balanced design in adversarial environments.

\vspace*{-0.5em}
\section{Conclusion}
In this letter, we have investigated a device-edge CI system under malicious jamming.
By applying data regression techniques, we have modeled inference accuracy as a function of SINR and the partitioning point, capturing the impacts of jamming and DNN partitioning.
On this basis, we have formulated a problem to maximize the system's RDA by jointly optimizing computation resource allocation, devices' transmit power, and DNN partitioning subject to inference accuracy requirements and computing resource constraints.
To tackle this MINLP problem, we have designed an efficient AO-based algorithm that decomposes the problem into three subproblems, each solved via the KKT conditions, convex optimization methods, and the QGA, respectively.
Simulation results have demonstrated that our proposed scheme outperforms all baselines.
Given the openness of wireless channels, future work will focus on jointly optimizing model partitioning and anti-jamming secure transmission strategies for CI.

\end{document}